\documentclass{ieeeaccess}
\usepackage{cite}
\usepackage{amsmath,amssymb,amsfonts}
\usepackage{graphicx}
\usepackage{textcomp}

\usepackage{multirow}
\usepackage{subfigure}

\usepackage{algorithm}  
\usepackage{algorithmicx}  
\usepackage{algpseudocode}  
\usepackage{amsmath}  
  
\floatname{algorithm}{Algorithm}

\def\BibTeX{{\rm B\kern-.05em{\sc i\kern-.025em b}\kern-.08em
    T\kern-.1667em\lower.7ex\hbox{E}\kern-.125emX}}
\begin{document}
\history{Date of publication xxxx 00, 0000, date of current version xxxx 00, 0000.}
\doi{10.1109/ACCESS.2017.DOI}

\bibliographystyle{plain}

\title{A New Data Normalization Method to Improve Dialogue Generation by Minimizing Long Tail Effect}
\author{\uppercase{Zhiqiang Zhan}\authorrefmark{1,2}
\uppercase{Zifeng Hou\authorrefmark{1,2}, and 
Yang Zhang}\authorrefmark{3}}
\address[1]{Institute of Computing Technology, Chinese Academy of Sciences, Beijing, China}
\address[2]{University of Chinese Academy of Sciences, Beijing, China}
\address[3]{Lenovo Research, Beijing, China}
%
%\markboth
%{Author \headeretal: Preparation of Papers for IEEE TRANSACTIONS and JOURNALS}
%{Author \headeretal: Preparation of Papers for IEEE TRANSACTIONS and JOURNALS}
%
%\corresp{Corresponding author: Zifeng Hou (e-mail: 1955houzifeng@sina.com); Yang Zhang (e-mail: zhangyang20@lenovo.com).}

\begin{abstract}
Recent neural models have shown significant progress in dialogue generation. 
Most generation models are based on language models. 
However, due to the Long Tail Phenomenon in linguistics, 
the trained models tend to generate words that appear frequently in training datasets, 
leading to a monotonous issue. 
To address this issue, we analyze a large corpus from Wikipedia 
and propose three frequency-based data normalization methods. 
We conduct extensive experiments based on transformers and three datasets 
respectively collected from social media, subtitles, and the industrial application. 
Experimental results demonstrate significant improvements in diversity and informativeness (\emph{defined as the numbers of nouns and verbs}) of generated responses. 
More specifically, the unigram and bigram diversity are increased by 2.6\%-12.6\% and 2.2\%-18.9\% on the three datasets, respectively. 
Moreover, the informativeness, i.e. the numbers of nouns and verbs, 
are increased by 4.0\%-7.0\% and 1.4\%-12.1\%, respectively. 
Additionally, the simplicity and effectiveness enable our methods to be adapted to different generation models 
without much extra computational cost. 
 
\end{abstract}


\titlepgskip=-15pt

\maketitle

\section{Introduction}\label{section:introduction}
With the benefit of Recurrent Neural Network (RNN), 
especially Long-Short Term Memory (LSTM)\cite{DBLP:journals/neco/HochreiterS97} 
and Gated Recurrent Unit (GRU)\cite{DBLP:journals/corr/ChungGCB14}, 
significant progress has been achieved in the sequence tasks, such as dialogue generation. 
As the key module of dialogue agents, 
dialogue generation models play an important role in facilitating interactions between humans and machines. 
Most recent dialogue generation models are based on the encoder-decoder framework. 
More specifically, in the framework, 
an encoder is used to generate an embedding representation of the input utterance, 
and then a decoder is adopted to generate a response 
based on the language model according to the embedding representation. 
As Seq2Seq models are able to offer the promise of scalability and language-independence, 
together with the capacity to implicitly learn semantic and syntactic relations between pairs, 
and to capture the contextual dependencies \cite{DBLP:conf/naacl/SordoniGABJMNGD15} 
in a way not possible with conventional Statistical Machine Translation (SMT) 
approaches\cite{DBLP:conf/emnlp/RitterCD11}, 
the most common models based on the encoder-decoder framework are Seq2Seq models, 
such as the works in \cite{DBLP:conf/nips/SutskeverVL14,
DBLP:journals/corr/VinyalsL15,
DBLP:conf/naacl/SordoniGABJMNGD15,
DBLP:conf/aaai/SerbanSBCP16}.

%\textcolor{blue}{As the common Long Tail Phenomenon exists in linguistics, 
%illustrated in Figure~\ref{fig:long-tail} and Figure~\ref{fig:long-tail-the}, 
%language models are vulnerable to this issue. 
%Consequently, dialogue generation models based on the language model 
%prefer to generate more frequent words, 
%which raises a problem of poor diversity}. 

\Figure[b!](topskip=0pt, botskip=0pt, midskip=0pt)[scale=0.4]{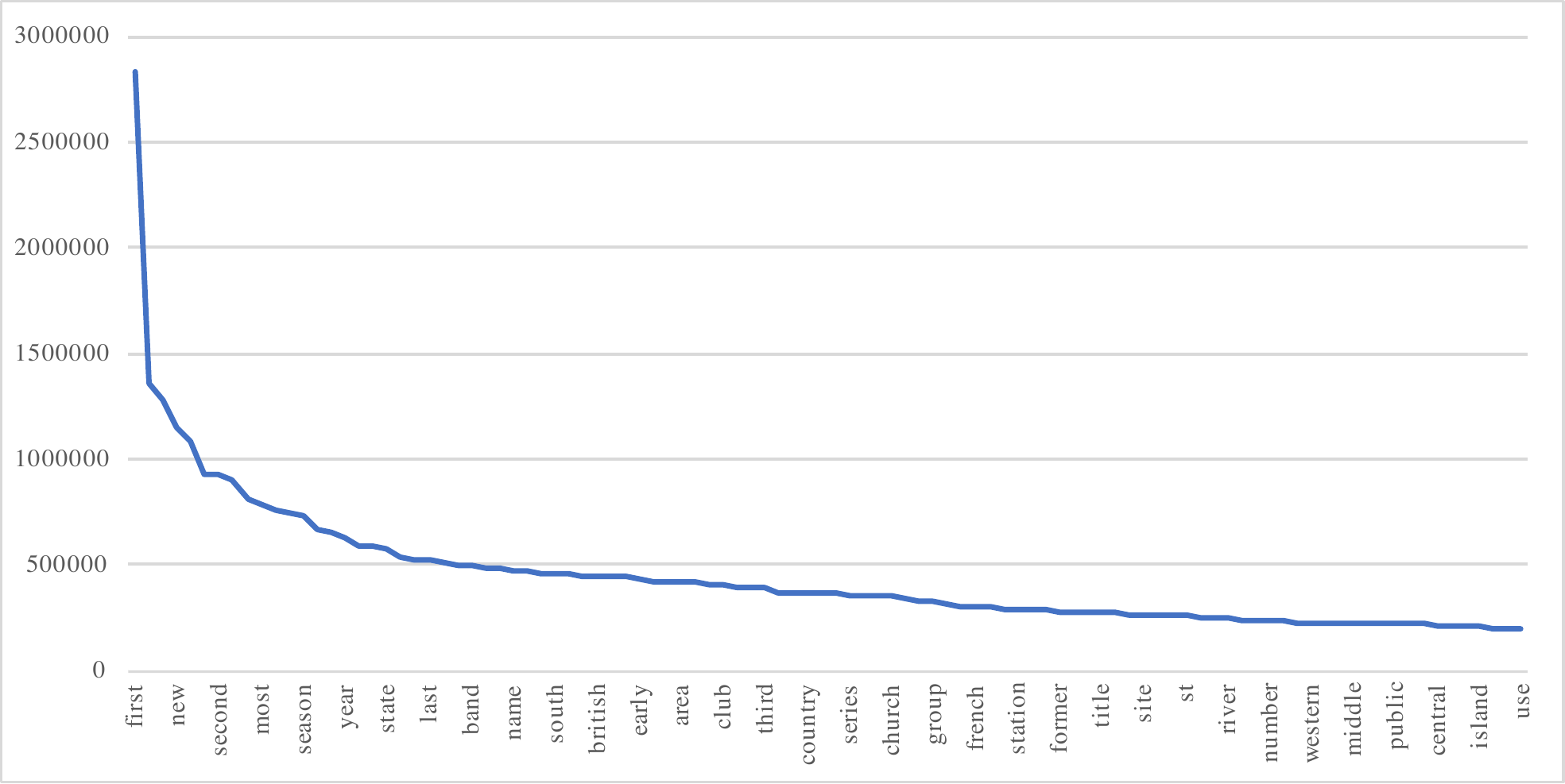}
{The frequency distribution of the top-200 most frequent bigrams beginning with \emph{``the''} in Wikipedia. \label{fig:long-tail-the}}

%\begin{figure}[h!]
%\centering
%\includegraphics[scale=0.4]{long-tail-all-words.pdf}
%\caption{The frequency distribution of the top-1000 most frequent words in Wikipedia. }
%\label{fig:long-tail}
%\end{figure}

\begin{figure}[t!]
\centering
\includegraphics[scale=0.63]{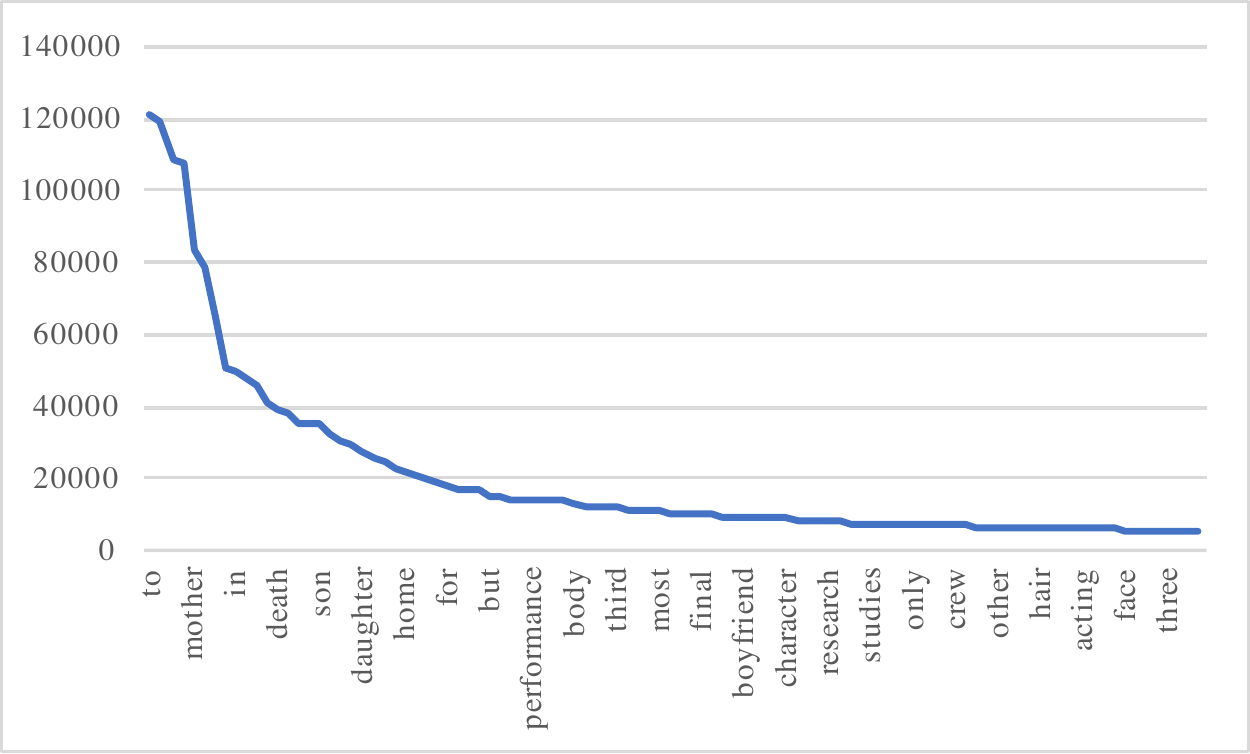}
\caption{The frequency distribution of the top-200 most frequent bigrams beginning with \emph{``her''} in Wikipedia. }
\label{fig:long-tail-her}
\end{figure}

An engaging dialogue generation model should be able to output grammatical, 
coherent responses that are diverse and informative. 
However, due to the common Long Tail Phenomenon in linguistics, 
illustrated in Figure~\ref{fig:long-tail-the} and Figure~\ref{fig:long-tail-her}, 
and the common training loss function \emph{Cross-Entropy}, 
neural conversation models tend to generate trivial or non-committal responses, 
often involving high-frequency phrases along the lines of \emph{``I don't know''} or \emph{``I'm OK''} \cite{DBLP:journals/corr/VinyalsL15,DBLP:conf/naacl/SordoniGABJMNGD15,DBLP:conf/aaai/SerbanSBCP16}. 
Consequently, dialogue generation models based on the language model 
are vulnerable to this issue, 
raising problems of poor diversity and less informativeness. 

In order to address this issue,  Li et al. propose a diversity-promoting objective function, 
i.e. Maximum Mutual Information (MMI)\cite{DBLP:conf/naacl/LiGBGD16}. 
MMI models consist of two models: one is trained to minimize $p(Y|X)$ and the other is trained to minimize $p(X|Y)$, 
specifically, $X$ and $Y$ are input utterances and responses respectively. 
During the decoding procedure, beam search is used to obtain better responses according to MMI. 
Shi et al. introduce Reinforcement Learning (RL) and propose Inverse Reinforcement Learning (IRL) framework 
to promote the diversity of dialogue generations\cite{DBLP:conf/ijcai/ShiCQH18}. 
The IRL framework learns a reward function on training data, 
and an optimal policy to maximize the expected total reward. 
The reward and policy functions in IRL are optimized alternately to generate higher quality texts.
Inspired by the work in \cite{DBLP:conf/nips/GoodfellowPMXWOCB14}, 
Xu et al. utilize Generative Adversarial Networks (GAN) to promote the diversity of dialogue generations\cite{DBLP:conf/emnlp/XuRL018}. 
Besides, Gao et al. propose a SPACEFUSION model to jointly optimize diversity and relevance\cite{DBLP:conf/naacl/GaoLZBGGD19}.
The model essentially fuses the latent space of a Seq2Seq model 
and that of an auto-encoder model by leveraging novel regularization terms.
In order to generate high-quality and informative conversation responses, 
Shao et al. propose a glimpse-model, which utilizes a stochastic beam-search algorithm with
segment-by-segment reranking to inject diversity earlier in the generation process\cite{DBLP:conf/emnlp/ShaoGBGSK17}. 
Zhou et al. propose an open-domain conversation generation model,  
which integrates commonsense knowledge into conversation generation, 
with the capacity for facilitating better generation 
through a dynamic graph attention mechanism\cite{DBLP:conf/ijcai/ZhouYHZXZ18}.

All the above methods of improving dialogue generation are 
either too complex and time-consuming to train the models or too inflexible to generalize the models.
In order to improve dialogue generation models, 
we propose three frequency-based data normalization methods, 
which are aimed at minimizing the Long Tail Effect during the training procedure 
to promote the performance of dialogue generation. 
With our proposed data normalization methods, 
the less frequent but valid words are able to obtain a larger generation possibility in the generation process. 
Thus, the generation models augmented with the proposed data normalization methods 
are able to promote the performance, especially the diversity. 
Moreover, the impressive advantage of the proposed methods is that 
they are independent of generation models, 
which makes it much more flexible to incorporate  them into various of dialogue generations. 
 
The paper is arranged as follows: We first introduce the related work in Section~\ref{section:related-work}. 
Then we introduce our motivation in Section~\ref{section:motivation}. 
Then we detail our proposed data normalization methods in Section~\ref{section:normalization-method}. 
Subsequently, we elaborate on our experiments in Section~\ref{section:experiments}, 
followed by the experimental results and thorough analysis in Section~\ref{section:res-analysis}. 
The last comes the conclusion part in Section~\ref{section:conclusion}. 

\section{Related Work}\label{section:related-work}
Recently, much work has been done in dialogue generation with the encoder-decoder architecture,
 which maps input utterances to the corresponding responses. 
 For example, Vinyals and Le build a neural conversational model based on the Seq2Seq architecture, 
with an encoder to encode the context and a decoder to generate a reply\cite{DBLP:journals/corr/VinyalsL15}. 
Serban et al. extend the hierarchical recurrent encoder-decoder neural network 
to the dialogue generation domain and achieve competitive performance on multi-turn dialogue generation\cite{DBLP:conf/aaai/SerbanSBCP16}. 

Although significant progress has been achieved in dialogue generation recently, 
there are still a few obstacles to further enhancement of generation performance. 
For example, due to the Long Tail Phenomenon in linguistics  
% which is illustrated in Figure~\ref{fig:long-tail} and Figure~\ref{fig:long-tail-the}. 
and the common training loss function \emph{Cross-Entropy}, 
language model-based dialogue generation models tend to generate frequent words in the training dataset, 
leading to a monotonous issue. 
The Long Tail Phenomenon is very common in many domains, 
such as recommendation and linguistics. 
Thus, it is essential to mitigate the Long Tail Effect. 
For example, Zhang and Luo propose Deep Neural Network structures serving as feature extractors 
to mitigate the Long Tail Effect on Twitter 
and solve hate detection task\cite{DBLP:journals/semweb/ZhangL19}. 
Hamedani and Kaedi utilize personalized diversification 
to recommend the long tail items\cite{DBLP:journals/kbs/HamedaniK19}.
Sreepada and Patra use few shot learning technique to weaken the Long Tail Effect 
in recommendations\cite{DBLP:journals/eswa/SreepadaP20}. 

In order to address the monotonous issue, 
Li et al. utilize Maximum Mutual Information (MMI) 
as the loss function instead of maximum likelihood\cite{DBLP:conf/naacl/LiGBGD16}. 
Although MMI models are able to promote the diversity of generations, 
with an extra model and the beam search procedure, 
MMI models are too time-consuming and resource-consuming in both training and testing procedures. 
Different from \cite{DBLP:conf/naacl/LiGBGD16}, 
Shi et al. adopt Reinforcement Learning to promote diversity\cite{DBLP:conf/ijcai/ShiCQH18}. 
In order to calculate the expected total reward, 
time-consuming MCMC sampling is used, 
leading to a much longer training procedure. 
Xu et al. first apply GAN to enhance the diversity of dialogue generations\cite{DBLP:conf/emnlp/XuRL018}. 
Because of the instability of training, 
GAN models are difficult to converge. 
Thus, much experience is needed to train GAN models. 
Shao et al. propose a glimpse-model to enhance the diversity of generation model\cite{DBLP:conf/emnlp/ShaoGBGSK17}. 
With a stochastic beam-search algorithm, 
the model is able to generate high-quality and informative conversation responses. 
However, the procedure of beam-search and reranking to inject diversity is much time-consuming. 
Zhou et al. propose an open-domain conversation model 
which incorporates commonsense knowledge in the model\cite{DBLP:conf/ijcai/ZhouYHZXZ18}. 
Although the model is able to facilitate better generation, 
a large commonsense knowledge base is essential and critical, 
which is difficult to be constructed. 
Gao et al. propose a SPACEFUSION model, 
which can jointly optimize diversity and relevance\cite{DBLP:conf/naacl/GaoLZBGGD19}. 
The model first obtains the latent space of $X$ and $Y$ through a Seq2Seq and an auto-encoder models respectively, 
$X$ and $Y$ are input utterances and responses respectively. 
Then the model fuses both latent spaces by leveraging novel regularization terms. 
However, due to the inherent recurrent procedure of the recurrent neural network adopted in Seq2Seq, 
the model is too time-consuming and source-consuming to handle long sentences. 

Different from all the above improvement methods and models, 
we propose three simple but effective frequency-based normalization methods, 
specifically, Log Normalization (LN), Mutual Normalization (MN) and Log Mutual Normalization (LMN), 
aiming to weaken the effect of Long Tail in linguistics, 
so as to improve dialogue generation models. 
At first, we analyze large corpus and obtain the distribution information of N-grams (in our experiments N=2). 
Then, we filter the top-K (in our experiments K=200) most frequent N-grams which are used in the decoder module of the dialogue generation models. 
In order to fully explore how to effectively weaken the Long Tail Effect, 
we design three data normalization methods to utilize the obtained N-gram distribution information. 
Benefiting from the high efficiency, 
the proposed data normalization methods barely increase computing complexity and resource consumption. 
Furthermore, as an independent module, 
our data normalization methods are flexible to be augmented with various dialogue generation models conveniently and directly. 
With the data normalization methods, 
the less frequent but sound words gain a larger generation possibility in the training procedure, 
which is able to foster the diversity of generation significantly. 

The most common architectures of recent dialogue generation models are based on reinforced RNNs, 
such as LSTM\cite{DBLP:journals/neco/HochreiterS97} and GRU\cite{DBLP:journals/corr/ChungGCB14}. 
Even though these models bring significant improvements in sequence tasks, 
the encoder consisting of RNN costs too much time at the training stage because of the recurrent procedure. 
In order to accelerate model training, 
many researchers explore means of replacing the recurrent procedure in RNN with the parallel procedure. 
Inspired by the significant success of the attention mechanism in computer vision\cite{DBLP:conf/icml/XuBKCCSZB15}, 
Bahdanau et al. introduce attention to Machine Translation (MT) task\cite{DBLP:journals/corr/BahdanauCB14}. 
Subsequently, Vaswani et al. take full advantage of attention mechanism 
and propose transformer architecture which enables the encoder procedure to run parallel\cite{DBLP:conf/nips/VaswaniSPUJGKP17}. 
With the benefit of efficiency and effectiveness, 
transformer architecture gains growing attention since its creation. 
For example, Shao et al. utilize a transformer-based model to solve Question Answering (QA) tasks\cite{DBLP:journals/access/ShaoGCH19}.
Devlin et al. propose BERT to deal with several classification tasks in Natural Language Processing (NLP) 
and achieve the-state-of-the-art in several tasks\cite{DBLP:conf/naacl/DevlinCLT19}. 
Therefore, our models are mainly based on the mainstream transformer architecture. 

\section{Motivation} \label{section:motivation}
The most common loss function used in dialogue generation model is cross-entropy, 
and the formula is as follows:
\begin{align}
Loss &= \frac{1}{M}  \sum_{i=1}^{M} \sum_{j=1}^{N} - p_{ij} *log(\hat{p}_{ij}) \\
\hat{p} &= p(Y|X )
\label{equation:loss_function}
\end{align}
where $M$ is the number of words in a sentence, $N$ is the size of the vocabulary, 
$p_{ij}$ is the true probability distribution for $i^{th}$ word, 
and $\hat{p}_{ij}$ is the model prediction probability distribution for $i^{th}$ word, 
$X$ is the input utterance, $Y$ is the corresponding response. 
The target of training model is to minimize $Loss$, 
which essentially makes model fit the probability distribution of the training dataset. 
As a consequence, the model prefers frequent words in the training dataset, 
leading to low diversity. 

In order to solve this issue to improve the performance of dialogue generation, 
we are inspired to improve the common loss function cross-entropy, 
aiming to balance the generation probability of frequent words and less frequent but sound words. 

Thus, we improve the loss function as follows:
\begin{align}
Loss &= \frac{1}{M}  \sum_{i=1}^{M} \sum_{j=1}^{N} - p_{ij} *log(p'_{ij}) \\
p'_{ij} &= Norm(\hat{p})	\label{equation:norm}
\end{align}
where $M$, $N$, $p_{ij}$ and $\hat{p}$ are the same with Equation~\ref{equation:loss_function}, 
$Norm$ is our proposed data normalization method, 
which is expatiated in Section~\ref{section:normalization-method}. 

At last, we minimize $Loss$ to train the model. 

\section{Data Normalization Methods}\label{section:normalization-method}

In order to explore a valid normalization method, 
we analyze a large corpus Wikipedia which is easy to acquire, 
and the detail information is listed in Table \ref{table:wiki-info}. 
After a thorough analysis, 
we propose three data normalization methods: 
Log Normalization (LN), Mutual Normalization (MN) and Log Mutual Normalization (LMN), 
which are elaborated in the following respectively. 
\begin{table}
\begin{center}
\begin{tabular}{|l|l|l|}
%\begin{tabular}{llllllllllll}
\hline
\bf Article Num & \bf Word Num & \bf File Size\\
\hline 
4.58 Million	&	2.56 Billion	&	15.83 GB\\
 \hline 
\end{tabular}
\end{center}
\label{table:wiki-info}
\caption{ The statistical information on Wikipedia corpus.}
\end{table}

\subsection{Log Normalization}
Due to the Long Tail Phenomenon in linguistics, 
given the previous word, 
the frequency of the next words differs with a large margin. 
For example, given the previous word \emph{``for''}, 
the top-10 of most frequent next words are \emph{``the''}(5,119,569), \emph{``his''}(537,910), \emph{``example''}(285,066), 
\emph{``its''}(224,253), \emph{``their''}(220,962),
\emph{``her''}(211,172), \emph{``best''} (195,793), \emph{``an''}(180,116), \emph{``this''}(159,481), and \emph{``all''} (126,973) respectively. 
Because of the large frequency margin between \emph{``the''} and \emph{``his''}, 
language models prefer to generate \emph{``Tom got a new table for \textbf{the} dining room.''} instead of 
\emph{``Tom got a new table for \textbf{his} dining room.''} even though both sentences sound good. 

In order to shrink the frequency margin, 
specifically, to penalize the high-frequency words, 
we intuitively take the logarithm on frequency first, 
then followed by a linear normalization, 
which is called Log Normalization (LN). 
The computational procedure is as follows: 
\begin{align}
prob &= \frac{log(Freq(\overline{w_a w_b}) + \varepsilon)}{Max_v}	\\
Max_v & = log(max_{i=0}^{k-1}{Freq(\overline{w_a w_i})}  + \varepsilon)
\end{align}
where $Freq$ is a map containing the frequency of unigrams and bigrams, 
$Freq(\overline{w_a w_b})$ is the frequency of bigram $\overline{w_a w_b}$, 
$w_i$ is a word following $w_a$, $\varepsilon$ is a smooth factor to avoid divide-by-zero. 

In order to detail the computational procedure of LN, 
we list the pseudo-code in Algorithm~\ref{algorithm:ln}. 
   \begin{algorithm} [htbp] 
        \caption{Log Normalization}  
        \begin{algorithmic}[1] %每行显示行号  
            \Require $Freq$ unigram and bigram count, $w_a$ previous word, $w_b$ current word 
            \Ensure prob  
            \Function {LogNormalization}{$Freq, w_a, w_b$}  
                \State $result \gets 0$  
                \State $Max_v \gets 0$  
                \State $\varepsilon \gets 1.1$
                \For{$i = 0 \to k-1$}  
                    \State $ Max_v \gets max(Max_v, Freq(\overline{w_a w_i}))$  
                \EndFor  
                \State $result \gets \frac{log(Freq(\overline{w_a w_b}) + \varepsilon)}{log(Max_v + \varepsilon)}$  
                
                \State \Return{$result$}  
            \EndFunction  
        \end{algorithmic}  
        \label{algorithm:ln}
    \end{algorithm}

\vspace{-0.1in}    

\subsection{Mutual Normalization}
Log Normalization takes a simple but effective way to minimize the effect of Long Tail in linguistics. 
However, it does not take mutual information of bigrams into account. 
Considering $Freq(\overline{w_a w_b})$ is partly related to $Freq(w_b)$ 
(large $Freq(w_b)$ probably indicating large $Freq(\overline{w_a w_b})$),  
we propose another normalization method to eliminate this factor. 
More specifically, similar to the mutual information, 
we divide the bigram frequency by the unigram frequency to shrink the frequency margin, 
which is called Mutual Normalization (MN). 
The computational procedure is as follows, 
and the pseudo-code is listed in Algorithm~\ref{algorithm:mn}: 
\begin{align}
prob = \frac{Freq(\overline{w_a w_b})}{Freq(w_b) + \varepsilon}
\end{align}
where $Freq$ is a map containing the frequency of unigrams and bigrams, 
$Freq(\overline{w_a w_b})$ is the frequency of bigram $\overline{w_a w_b}$, 
$Freq(w_b)$ is the frequency of unigram $w_b$,
 $\varepsilon$ is a smooth factor to avoid divide-by-zero. 
    \begin{algorithm}  [htbp]
        \caption{Mutual Normalization}  
        \begin{algorithmic}[1] %每行显示行号  
            \Require $Freq$ unigram and bigram count, $w_a$ previous word, $w_b$ current word 
            \Ensure prob  
            \Function {MutualNormalization}{$Freq, w_a, w_b$}  
                \State $result \gets 0$ 
                \State $\varepsilon \gets 0.1$
                \State $result \gets \frac{Freq(\overline{w_a w_b})}{Freq(w_b) + \varepsilon}$   
                \State \Return{$result$}  
            \EndFunction  
        \end{algorithmic}  
        \label{algorithm:mn}
    \end{algorithm}

\vspace{-0.1in}    
\subsection{Log Mutual Normalization}
At last, we combine the advantages of Log Normalization and Mutual Normalization 
to propose a third normalization method. 
More specifically, we first take a logarithm on both bigram and unigram frequency, 
which makes the probability distribution smoother. 
Then we do a mutual information normalization, 
which is able to eliminate the affect of frequency of the posterior words 
to calculate the probability of bigrams. 
Thus, this normalization method is called Log Mutual Normalization (LMN). 
The computational procedure is as follows, 
and we list the pseudo-code in Algorithm~\ref{algorithm:lmn}: 
\begin{align}
prob = \frac{log(Freq(\overline{w_a w_b}) + \varepsilon)}{log(Freq(w_b) + \varepsilon)}
\end{align}
where $Freq$ is a map containing the frequency of unigrams and bigrams, 
$Freq(\overline{w_a w_b})$ is the frequency of bigram $\overline{w_aw_b}$, 
$Freq(w_b)$ is the frequency of unigram $w_b$, 
$\varepsilon$ is a smooth factor to avoid divide-by-zero. 

    \begin{algorithm} [htbp] 
        \caption{Log Mutual Normalization}  
        \begin{algorithmic}[1] %每行显示行号  
            \Require $Freq$ unigram and bigram count, $w_a$ previous word, $w_b$ current word 
            \Ensure prob  
            \Function {LMNormalization}{$Freq, w_a, w_b$}  
                \State $result \gets 0$  
                \State $\varepsilon \gets 1.1$
                \State $result \gets \frac{log(Freq(\overline{w_a w_b}) + \varepsilon)}{log(Freq(w_b) + \varepsilon)}$  
                \State \Return{$result$}  
            \EndFunction  
        \end{algorithmic} 
        \label{algorithm:lmn} 
    \end{algorithm}

\begin{figure*}[t]
\vspace{-0.0in}
\begin{center}
\subfigure[The distribution of top-500 most frequent bigrams beginning with \emph{``for''}.]{
\label{fig:normed-for}
\includegraphics[scale=0.25, width=0.45\textwidth]{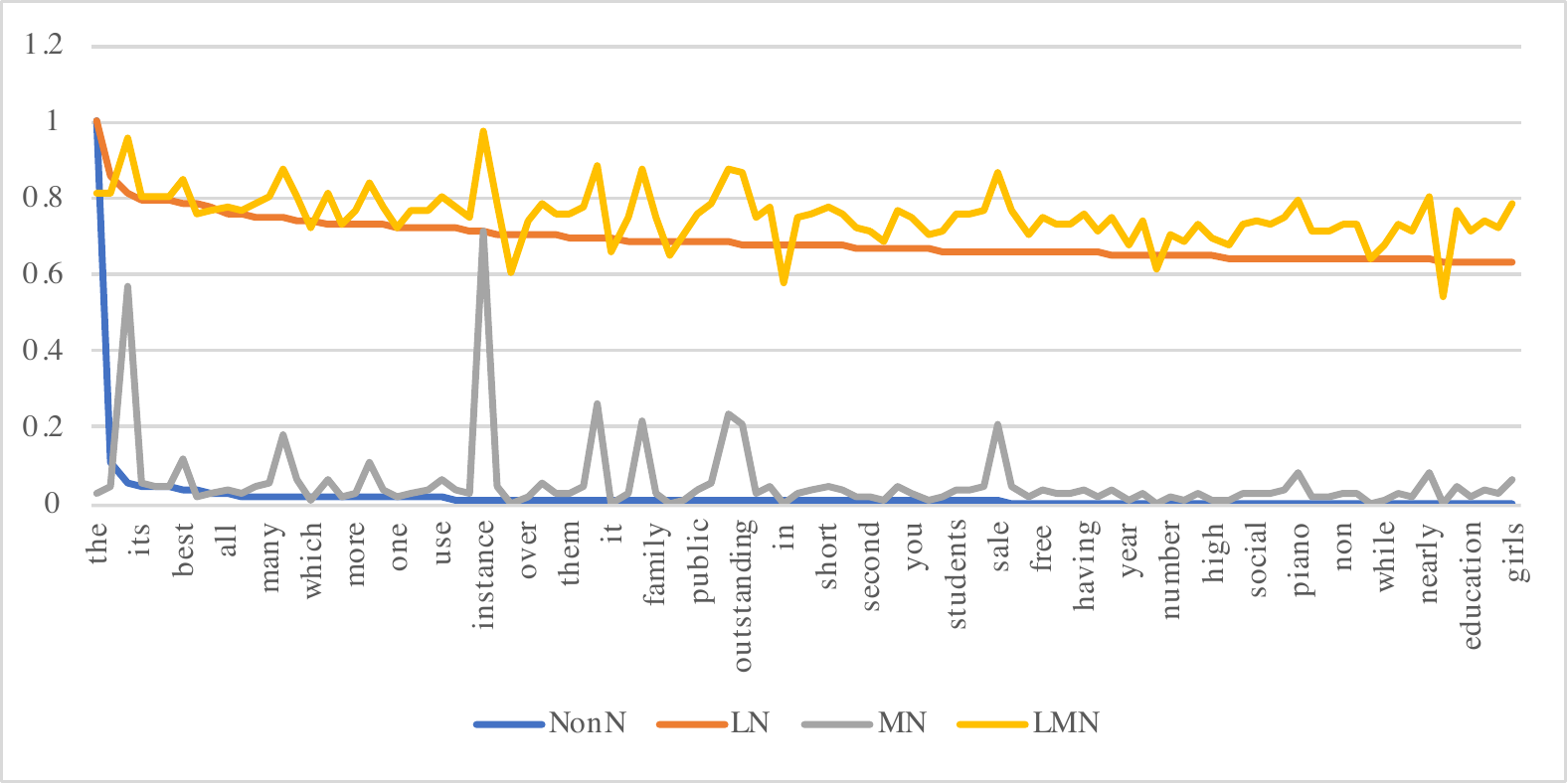}}
\hspace{0.1in}
\subfigure[The distribution of top-500 most frequent bigrams beginning with \emph{``would''}.]{
\label{fig:normed-would}
\includegraphics[scale=0.25, width=0.45\textwidth]{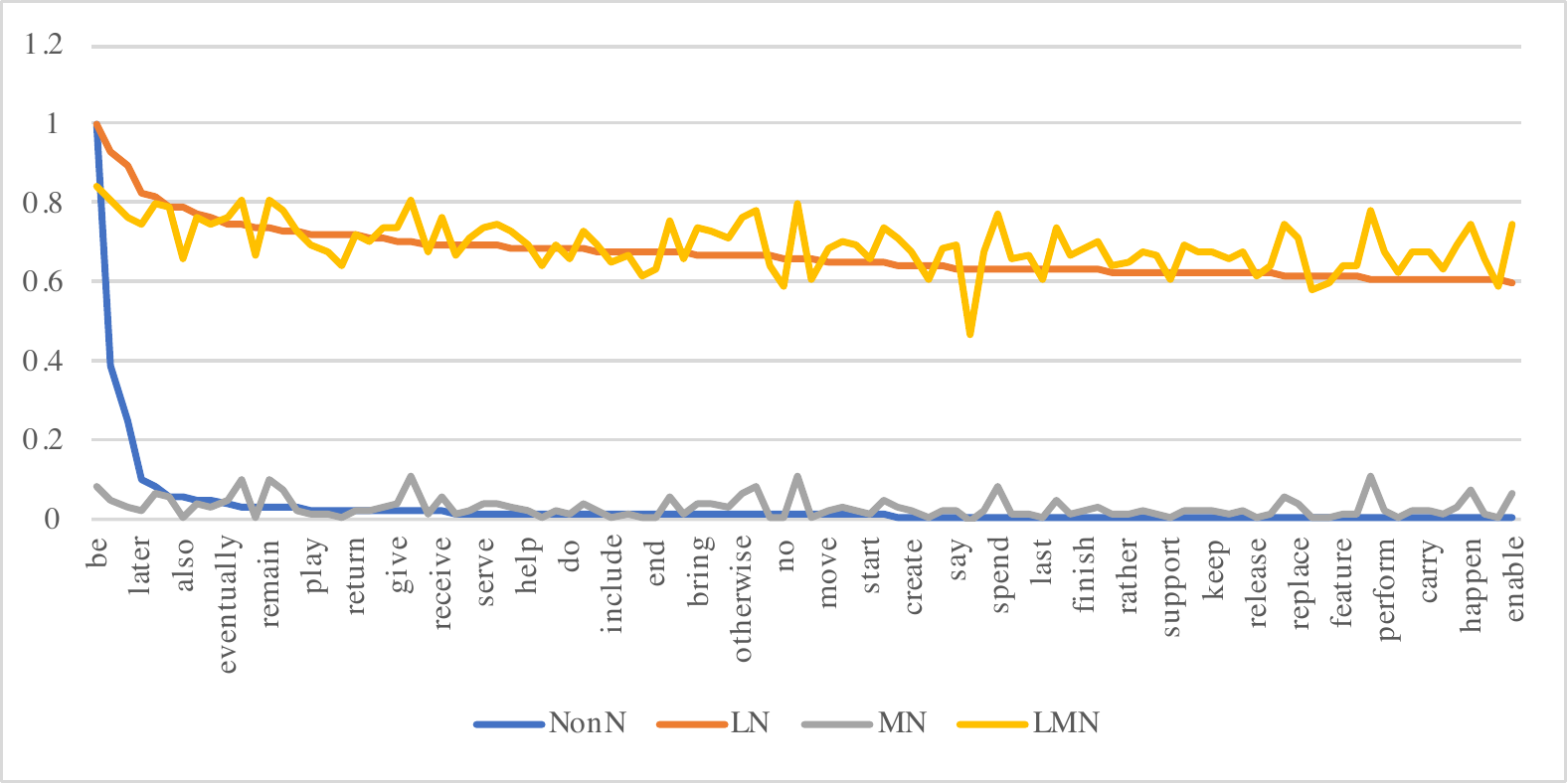}}
\vspace{-0.0in}
\subfigure[The distribution of top-500 most frequent bigrams beginning with \emph{``her''}.]{
\label{fig:normed-her}
\includegraphics[scale=0.25, width=0.45\textwidth]{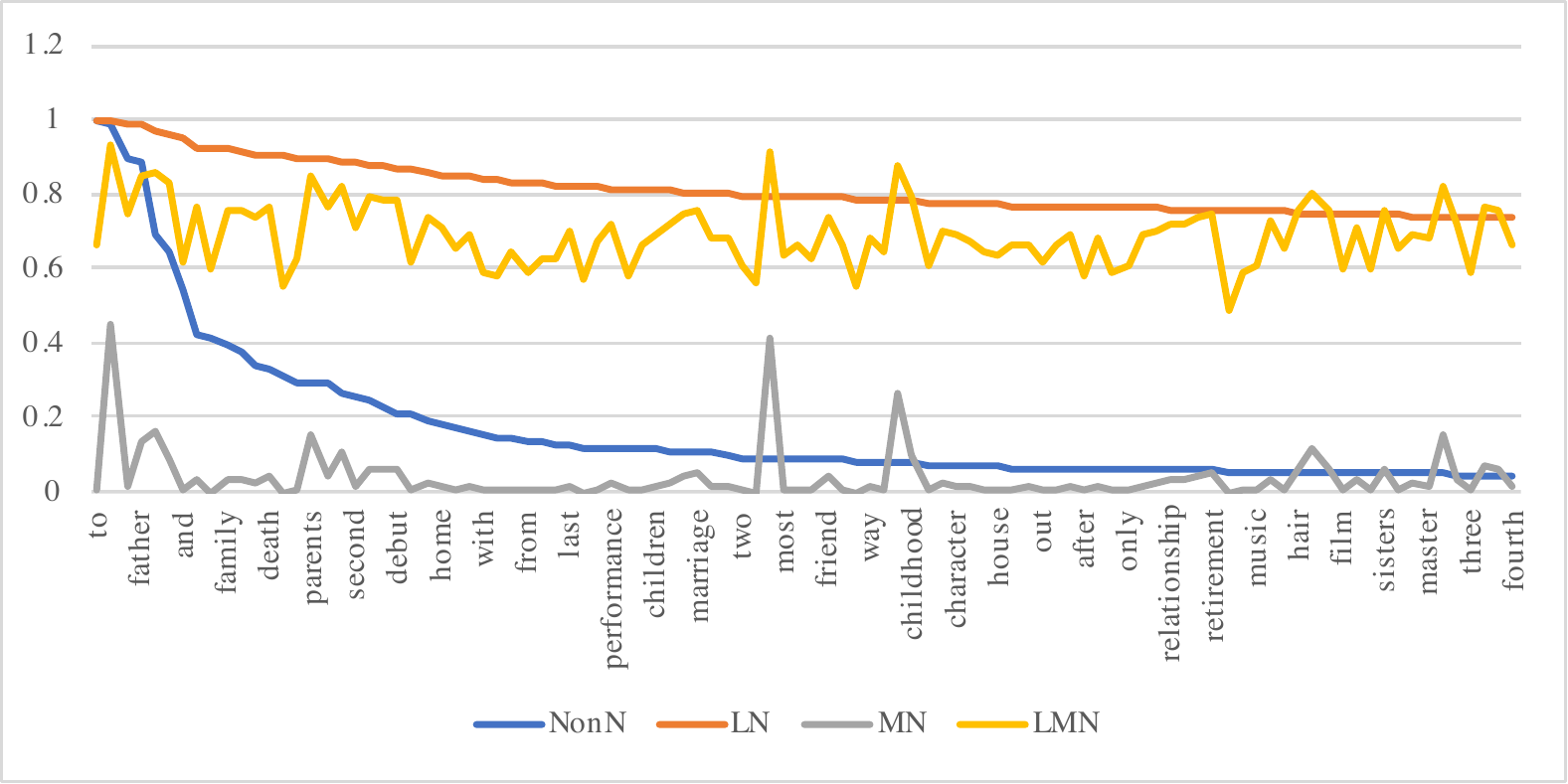}}
\hspace{0.1in}
\subfigure[The distribution of top-500 most frequent bigrams beginning with \emph{``win''}.]{
\label{fig:normed-win}
\includegraphics[scale=0.25, width=0.45\textwidth]{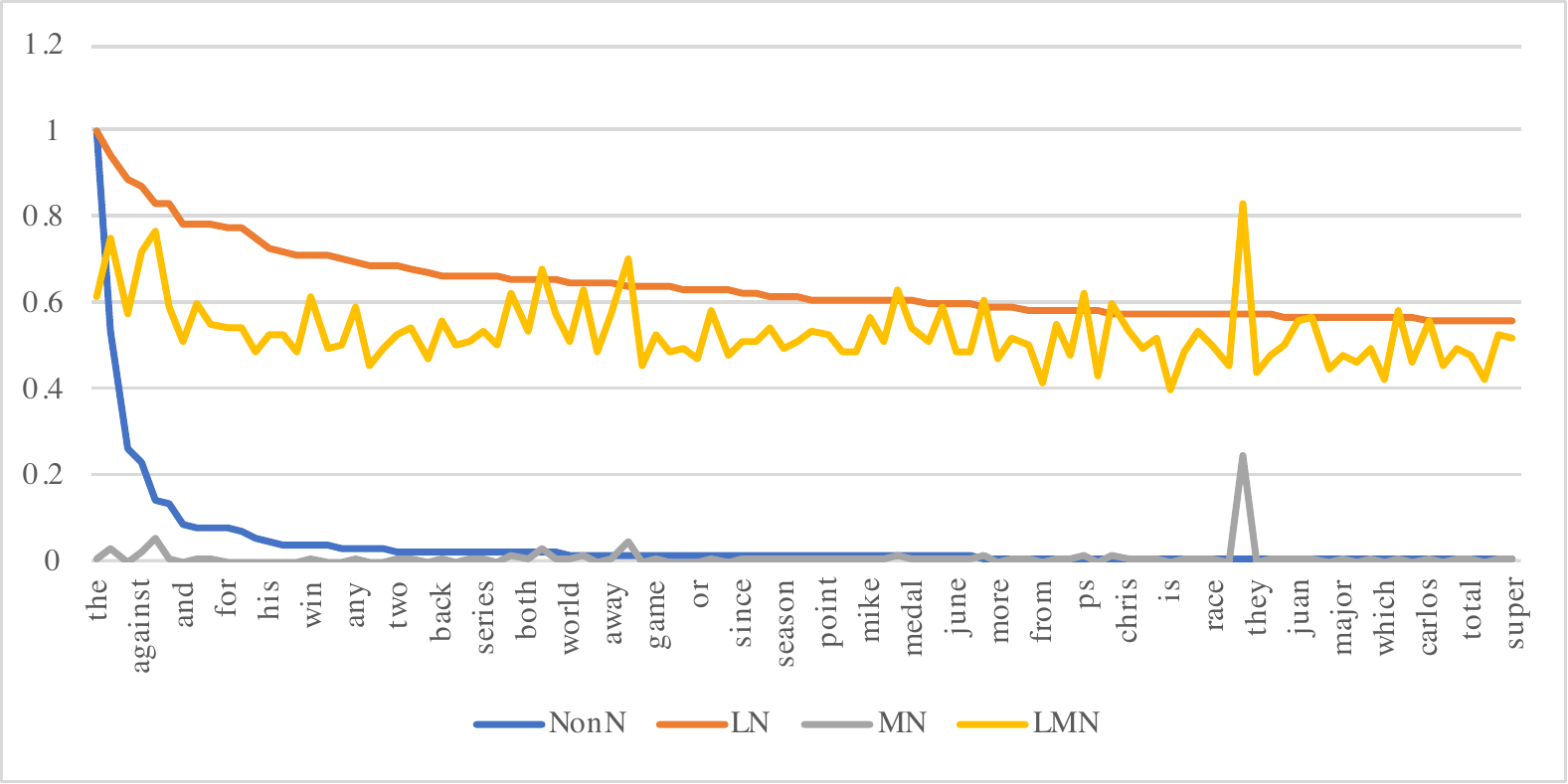}}
%\vspace{-0.15in}
\end{center}
\caption{The distribution of original and normalized bigram frequency of some bigrams.}
\label{fig:normed-sample}
\end{figure*}

In order to preliminarily validate the effectiveness of the above three normalization methods, 
we utilize them to normalize the bigram frequency of Wikipedia corpus. 
We give examples of words with different types (including preposition, pronoun, verb and modal verb), 
illustrated in Figure~\ref{fig:normed-sample}, 
where NonN, LN, MN, LMN indicate 
Non-Normalization, Log Normalization, Mutual Normalization,  
and Log Mutual Normalization respectively.

 According to Figure~\ref{fig:normed-sample}, 
with normalization methods, 
the margins of normalized frequencies are indeed shrunk significantly 
for different bigram types, including verb phrase, preposition phrase and noun phrase. 

Furthermore, we further analyze the mean and standard deviation values 
of the distribution of top-500 most frequent bigrams, 
which is listed in Table~\ref{table:normed-info}, 
where NonN, LN, MN, and LMN (\emph{keep the same in the rest}) indicate 
Non-Normalization, Log Normalization, Mutual Normalization, and 
Log Mutual Normalization respectively, 
mean indicates the average value of normalized frequency of bigrams, 
stddev indicates the standard deviation value of normalized frequency of bigram. 

\begin{table}[htbp]
\begin{center}
\begin{tabular}{|c|c|c|c|c|}
%\begin{tabular}{llllllllllll}
\hline
 & \bf NonN  & \bf LN & \bf MN & \bf LMN	\\ \hline 
\bf mean & 0.169	 & 0.813	& 0.038	& 0.691	\\ \hline
\bf stddev & 0.203	&	0.069	& 0.071	& 0.085	\\ \hline 
\end{tabular}
\end{center}
\caption{ The mean and standard deviation values of bigrams beginning with \emph{``her''} normalized by different methods. }
\label{table:normed-info}
%\vspace{-0.1in}
\end{table}

Compared with the non-normalization, 
the mean value becomes larger with a sharp margin for both Log Normalization
 and Log Mutual Normalization. 
Moreover, the standard deviation becomes much smaller for all three proposed normalization methods. 
The results indicate that 
the normalized probability distributions are more uniform 
and proposed normalization methods are able to reduce the effect of Long Tail Effectively. 

\section{Experimental Setups} \label{section:experiments}
\subsection{Datasets}
To examine the effectiveness of our normalization methods, 
we conduct a series of experiments on three datasets: 
Twitter, Subtitle, and Lenovo. 
The detailed information of these datasets is listed in Table~\ref{table:datasets}.

\begin{table*}
\begin{center}
\begin{tabular}{|l|l|l|l|l|l|l|l|}
%\begin{tabular}{llllllllllll}
\hline 
\bf  & \bf Dataset & \bf Voc Size & \bf Word Num & \bf UNK Num & \bf Sen Num & \bf Avg Len & \bf Word Ent\\
\hline 
\multirow{3}{*}{Train} & Twitter & 20,000 & 11,029,071 & 616,250 & 750,000 & 14.705 & 6.353 \\
& Subtitle & 15,753 & 5,803,971 & 246,145 & 440,000 & 13.191 & 5.862 \\
& Lenovo & 6,194 & 4,602,864 & 41,088 & 200,000 & 23.014 & 5.921 \\
\hline
\multirow{3}{*}{Test}& Twitter & 5,445 & 66,559 & 3,620 & 4,530 & 14.693 & 6.240 \\
 & Subtitle & 2,951 & 42,578 & 2,244 & 3,232 & 13.174 & 5.649 \\
 & Lenovo & 4,978 & 172,555 & 1,886 & 7,588 & 22.741 & 5.861 \\
 \hline 
\end{tabular}
\end{center}
\caption{ The statistical information of datasets. {Voc Size} is the number of distinct words, {Word Num} is the number of total words, {UNK Num} is the number of unknown words which is out of vocabulary, {Sen Num} is the number of all sentences, {Avg Len} is the average length of all sentences and {Word Ent} is the information entropy of all words.}
\label{table:datasets}
\end{table*}

Twitter dataset is a subset of the Twitter Corpus used in \cite{DBLP:conf/emnlp/RitterCD11}. 
Subtitle dataset is produced from the OpenSubtitles dataset used in \cite{DBLP:conf/naacl/LiGBGD16}. 
Lenovo dataset is composed of collected conversations between customers and online customer service. 
All the datasets used are keep the same with \cite{DBLP:conf/ijcnn/ZhanHYZZH19}. 
\subsection{Setups}
\begin{figure}
\centering
\includegraphics[scale=0.4]{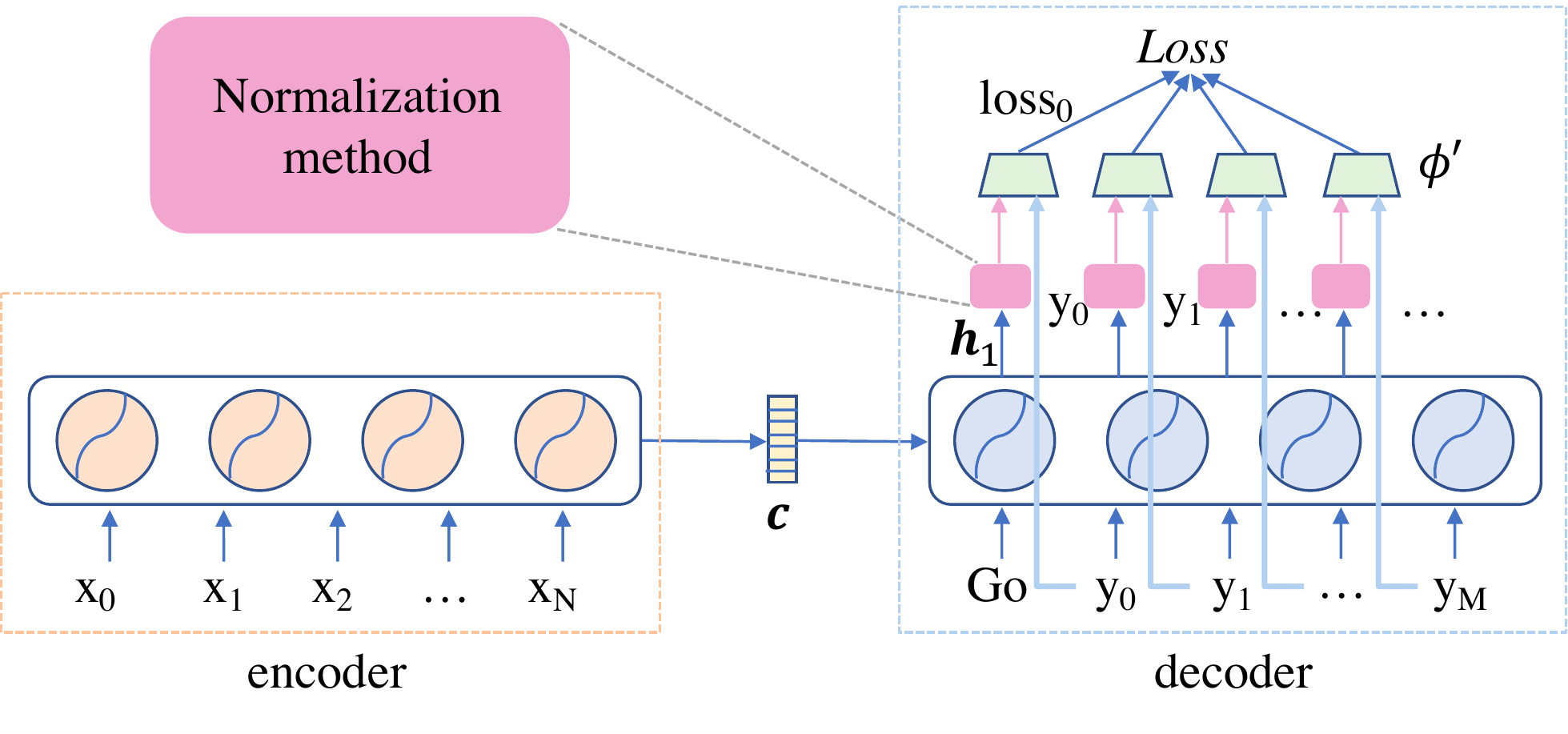}
\caption{The structure of the transformer augmented with data normalization methods. }
\label{fig:model}
\end{figure}

Because of the efficiency and effectiveness of transformer architecture, 
we conduct experiments based on the transformer. 
Contrast with the original transformer, 
we fuse it with proposed data normalization methods respectively 
to weaken the effect of Long Tail. 
Our model's structure is illustrated in Figure~\ref{fig:model}. 

The detailed setups of the model are listed as follows:
\begin{itemize}
\item The values of N (N-grams) and K(top-K) are set 2 and 200 respectively.  
\item The framework of training model is based on transformer. 
\item We filter the uncommon words in the training dataset with a threshold of value 10, 
i.e. words with a frequency less than 10 will be replaced by \emph{``$<$UNK$>$''} 
which indicates uncommon words. 
\item The max size of the vocabulary is set to 20000. 
As a consequence, The vocabulary sizes of Twitter, Subtitle, and Lenovo are 20000, 15753 and 6194 respectively. 
\item The gradient-based optimizer Adam \cite{DBLP:journals/corr/KingmaB14} is adopted to minimize the loss in training procedure for all datasets. 
\item The learning rate is set to 0.0001 for all datasets. 
\item The batch size is set to 256 for all datasets. 
\item The training epochs are set to 200, 200 and 100 for Twitter, Subtitle, and Lenovo respectively without early stop. 
\item The number of blocks in both encoder and decoder is set to 6. 
\item The dropout is only used in the embedding layer with a rate of 0.1. 
\end{itemize}

\section{Results and Analysis} \label{section:res-analysis} 
In order to validate our normalization methods, 
we conduct extensive experiments on the above three datasets. 
The experimental results are shown in Table~\ref{table:diversity}-\ref{table:dist-sen}.

\subsection{Experimental Results}
In order to check the diversity promoted, 
we analyze the results at different levels: word level, phrase level and sentence level. 
we calculate the number of distinct unigrams (i.e. at word level) 
and the number of distinct bigrams (i.e. at phrase level) respectively in the whole generated corpus, 
illustrated in Table~\ref{table:diversity}. 
Similarly, we calculate the number of distinct sentences (i.e. at sentence level) to analyze 
the sentence-level diversity promotion on Lenovo dataset, illustrated in Table~\ref{table:dist-sen}. 

In addition, we also evaluate the informativeness promotion with different normalization methods. 
Considering some words do not have much useful information, 
such as prepositions, conjunctions, and auxiliaries, 
we define the numbers of nouns and verbs as the informativeness of a sentence. 
Except for nouns and verbs, 
the average sentence length also partly indicates the informativeness of a sentence\cite{DBLP:conf/emnlp/ShaoGBGSK17}. 
Thus, we also take the average sentence length into account to analyze the informativeness promotion,
shown in Table~\ref{table:informativeness}. 
\begin{table}
\begin{center}
\begin{tabular}{|l|l|l|l|}
%\begin{tabular}{lllllllll}
\hline 
\bf Dataset & \bf Norm Method & \bf Unigram & \bf Bigram \\
\hline 
\multirow{4}{*}{Twitter} 
& NonN & 624 & 3,239 \\
& LN & 575 & 3,284 \\
& MN & 623 & \bf 3,310 (+2.2\%) \\
& LMN & \bf 640 (+2.6\%) & 3,131 \\
\hline
\multirow{4}{*}{Subtitle}
& NonN & 1,290 & 803 \\
& LN & 1,333 & 902 \\
& MN & 1,332 & 884 \\
& LMN & \bf 1,452 (+12.6\%) & \bf 955 (+18.9\%) \\
\hline
 \multirow{4}{*}{Lenovo}
& NonN & 510 & 1,190 \\
& LN & 485 & 1,221 \\
& MN & 500 & 1,237 \\
& LMN & \bf 525 (+2.9\%) & \bf 1,262 (+6.1\%) \\
 \hline 
\end{tabular}
\end{center}
\caption{ The word-level and phrase-level diversity of different normalization methods on three datasets. 
\emph{Norm Method} indicates the normalization method. 
\emph{Unigram} indicates the number of distinct unigrams, 
i.e. the size of vocabulary about the whole generated corpus. 
\emph{Bigram} indicates the number of distinct bigrams. 
Both of \emph{Unigram} and \emph{Bigram} are filtered to omit the uncommon words. }
\label{table:diversity}
\end{table}

\begin{table*}[bhtp]
\begin{center}
\begin{tabular}{|l|l|l|l|l|l|}
%\begin{tabular}{llllllllllll}
\hline 
 \bf Norm Method & \bf Distinct Sentence Num & \bf Top-5 Frequent Sentence  & \bf Sentence Num & \bf Top-5 Sentence Num \\ \hline 
 \multirow{5}{*}{NonN} & \multirow{5}{*}{2,557}  
 & \emph{``ok''}  & 362 & \multirow{5}{*}{687} \\
 & & \emph{``yes''} & 144 & \\ 
 & & \emph{``Yes''} & 73 & \\
 & & \emph{``Hi''} & 63 & \\
 & & \emph{``$<$NUMBER$>$''} & 45 & \\ \hline
  \multirow{5}{*}{LN} & \multirow{5}{*}{2,553 (-0.2\%)}  
 & \emph{``ok''} & 204 & \multirow{5}{*}{561 (-18.3\%)}  \\
 & & \emph{``yes''} & 174 & \\ 
 & & \emph{``Hi''} & 72 & \\
 & & \emph{``Yes''} & 56 & \\
 & & \emph{``$<$NUMBER$>$''} & 55 & \\ \hline
  \multirow{5}{*}{MN} & \multirow{5}{*}{\bf 2,642 (+3.3\%)}  
 & \emph{``ok''} & 230 & \multirow{5}{*}{442 (-35.7\%)}  \\
 & & \emph{``yes''} & 107 & \\ 
 & & \emph{``okay''} & 43 & \\
  & & \emph{``$<$NUMBER$>$''} & 34 & \\
  & & \emph{``No, I think that is all. Thank you very much!''}  & 28 & \\ \hline
  \multirow{5}{*}{LMN} & \multirow{5}{*}{2,631 (+2.9\%)}  
 & \emph{``ok''} & 196 & \multirow{5}{*}{\bf 426 (-38.0\%)}  \\
 & & \emph{``yes''} & 112 & \\ 
 & & \emph{``Yes''} & 53 & \\
 & & \emph{``$<$NUMBER$>$''} & 35 & \\
  & & \emph{``Hello!''} & 30 & \\ \hline
  \end{tabular}
\end{center}
\caption{The sentence-level diversity on Lenovo dataset. 
\emph{Norm Method} indicates the normalization method. 
\emph{Distinct Sentence Num} indicates the total number of distinct generated sentences, the larger the better. 
\emph{Total Sentence Num} indicates the total sentences of the top-5 most frequent generated sentences, the smaller the better. 
\emph{Top-5 Frequent Sentence} indicates the  top-5 most frequent generated sentences and 
\emph{Sentence Num} indicates their frequency, the smaller the better. 
} 
\label{table:dist-sen}
\end{table*}

\begin{table*}[bhtp]
\begin{center}
\begin{tabular}{|l|l|l|l|l|l|l|}
%\begin{tabular}{llllll}
\hline 
\bf Dataset & \bf Norm Method & \bf NN Words & \bf VB Words & \bf Entity Words & \bf Avg Len \\
\hline 
\multirow{4}{*}{Twitter} 
 & NonN & 5,518 & 2,598 & 8,116 & \bf 11.7 \\
& LN & 5,437 &  \bf 2,634 (+1.4\%) & 8,071 & \bf 11.7 (+0\%) \\
& MN & 5,634 & 2,401 & 8,035 & 11.5 \\
& LMN & \bf 5,888 (+6.7\%) & 2,534 & \bf 8422 (+3.8\%) & 11.6 \\
\hline
\multirow{4}{*}{Subtitle}
& NonN & 2,492 & 1,365 & 3,857 & 8.5 \\
& LN & 2,388 & \bf 1,530 (+12.1\%) & 3,918 & 8.7 \\
& MN & 2,542 & 1,460 & 4,002 & 8.9 \\
& LMN & \bf 2,666 (+7.0\%) & 1,480 & \bf 4,146 (+7.5\%) & \bf 9.4 (+10.6\%) \\
\hline
 \multirow{4}{*}{Lenovo}
& NonN & 7,385 & 3,129 & 10,514 & 10.9 \\
& LN & 7,138 & 3,031 & 10,169 & 11.0 \\
& MN & 7,572 & \bf 3,313 (+5.9\%) & \bf 10,885 (+3.5\%) & \bf 11.5 (+5.5\%) \\
& LMN & \bf 7,679 (+4.0\%) & 3,185 & 10,864 & 11.2 \\
 \hline 
\end{tabular}
\end{center}
\caption{ The informativeness of different normalization methods on three datasets. 
\emph{Norm Method} indicates the normalization method. 
\emph{NN Words} indicates the number of nouns parsed from NLTK. 
\emph{VB Words} indicates the number of verbs parsed from NLTK. 
\emph{Avg Len} indicates the average length of sentences
Both of nouns and verbs are treated as \emph{Entity Words} indicating informativeness of a sentence.}
\label{table:informativeness}
\end{table*}

\begin{table}[htbp]
\begin{center}
\begin{tabular}{|l|l|l|l|}
%\begin{tabular}{llllllllllll}
\hline 
\bf Dataset & \bf Norm Method & \bf BLEU & \bf Entropy \\ \hline 
\multirow{4}{*}{Twitter} 
& NonN & 0.244 & 4.950 \\
& LN & \bf 0.276 (+13.1\%) &  4.914 \\
& MN & 0.265 & 4.982 \\
& LMN & 0.140 & \bf 4.896 (-1.1\%) \\
\hline
\multirow{4}{*}{Subtitle}
& NonN & 0.144 & 5.337 \\
& LN & 0.177 & \bf 5.335 (-0.04\%) \\
& MN & \bf 0.186 (+29.2\%) & 5.342 \\
& LMN & 0.183  & 5.408 \\
\hline
 \multirow{4}{*}{Lenovo}
& NonN & 0.789 & 5.050 \\
& LN & 0.822 & \bf 4.910 (-2.8\%) \\
& MN & \bf 0.882 (+11.8\%) & 4.996 \\
& LMN & 0.847 & 5.028 \\
 \hline 
\end{tabular}
\end{center}
\caption{ The performance of the model augmented with different normalization methods on three datasets. 
\emph{Norm Method} indicates the normalization method. 
\emph{BLEU} indicates the BLEU value calculated through NLTK, the larger the better. 
\emph{Entropy} indicates the information entropy of total words generated, the smaller the better. }
\label{table:bleu}
\end{table}

\begin{table}[htbp]
\vspace{-0.15in}
\begin{center}
\begin{tabular}{|l|l|l|l|l|}
%\begin{tabular}{llllllllllll}
\hline 
 \bf Word & \bf NonN & \bf LN & \bf MN & \bf LMN \\
\hline 
, & 19 & \bf 25 & 22 & 18 \\ \hline 
. & 44 & 39 & \bf 46 & 41 \\ \hline 
a & 21 & 12 & 16 & \bf 24 \\ \hline 
am & 3 & 5 & 4 & \bf 6 \\ \hline 
at & 2 & 2 & 2 & \bf 4 \\ \hline 
do & 9 & 9 & 7 & \bf 11 \\ \hline 
for & \bf 9 & 8 & 7 & \bf 9 \\ \hline 
have & 12 & \bf 15 & 12 & 12 \\ \hline 
I & 38 & 32 & \bf 42 & 34 \\ \hline 
my & 4 & \bf 6 & 2 & 3 \\ \hline 
so & 6 & 2 & 5 & \bf 8 \\ \hline
to & 26 & 30 & \bf 34 & 31 \\ \hline
what & 3 & 4 & 4 & \bf 6 \\ \hline
will & 5 & 4 & 5 & \bf 6 \\ \hline
with & 4 & \bf 5 & 4 & 3 \\ \hline
you & 10 & \bf 13 & \bf 13 & 12 \\  \hline
\bf SUM & 215 & 211 & 225 & \bf 228 \\ \hline

\end{tabular}
\end{center}
\caption{Some examples about the distribution of bigrams on Lenovo dataset. 
\emph{Word} indicates the first word of the bigram. 
The numbers in the table indicate the frequency of bigram generated from different normalization methods. 
The last row is the sum value of the above rows. 
} 
\label{table:ngram-sample}
\end{table}

Furthermore, we compare the BLEU \cite{DBLP:conf/acl/PapineniRWZ02} 
and information entropy \cite{DBLP:journals/sigmobile/Shannon01} of total words generated, 
shown in Table~\ref{table:bleu}. 
BLEU is usually used to automatically evaluate dialogue generation. 
Information entropy indicates the uncertainty of information. 
When choosing words randomly to generate a response, 
the model will get a uniform distribution of words, leading to high information entropy. 
However, due to the expression habit, the use frequency of each word is different, 
leading to low information entropy\cite{DBLP:conf/ijcnn/ZhanHYZZH19}. 
Thus, information entropy of total words partly indicates the performance of generation models. 

At last, we give examples of the distribution of some bigrams 
generated by different data normalization methods on Lenovo dataset, 
shown in Table~\ref{table:ngram-sample}. 

\subsection{Qualitative Analysis}
According to the above experimental results, 
we conduct a thorough analysis from four aspects: 
\emph{Word-level and Phrase-level Diversity Promotion}, \emph{Sentence-level Diversity Promotion}, 
\emph{Informativeness Promotion}, and\emph{Performance Promotion}. 

\subsubsection{Word-level and Phrase-level Diversity Promotion}
To validate our normalization methods, 
we first compare the diversity of unigrams and bigrams generated with different normalization methods, 
which is shown in Table~\ref{table:diversity}. 
The results clearly indicate that the performance is enhanced by data normalization methods, 
especially on Subtitle dataset with an improvement up to 18.9\% on bigram diversity. 
The main cause is that our normalization methods effectively reduce the effect of Long Tail in linguistics 
by penalize the frequent words, 
resulting in a promotion of suitable but less frequent words, which is illustrated in Figure~\ref{fig:normed-sample}. 
Comparing with Log Normalization, Mutual Normalization and Log Mutual Normalization get better performance, 
because Log Normalization only shrinks the margin of words' frequency 
so that the most frequent word still keeps the largest probability. 
However, benefiting from the affect of mutual information, 
both Mutual Normalization and Log Mutual Normalization are able to change the order of words' frequency. 
As a consequence, the performance is promoted with a larger scale. 
Combining the advantages of both Log Normalization and Mutual Normalization, 
Log Mutual Normalization almost gets the best performance 
on both word-level and phrase-level diversity for all the three datasets.   

\subsubsection{Sentence-level Diversity Promotion}
In addition, we evaluate the diversity promotion at sentence level. 
We analyze the top-5 of most frequent generated sentences and their total number, 
which is shown in Table~\ref{table:dist-sen}. 
According to the content of Table~\ref{table:dist-sen}, 
the model augmented with Mutual Normalization generates more distinct sentences on Lenovo dataset. 
Furthermore, it is clear that the most frequent generated sentences are uninformative, 
such as \emph{``ok''}, \emph{``yes''} and \emph{``Hi''}. 
Without data normalization methods, 
the model generates more uninformative sentences up to 18.1\%. 
However, with Log Mutual Normalization, 
the uninformative sentences generated are significantly reduced by 38.0\%. 
The results verify that our data normalization methods are indeed able to reduce uninformative words 
and promote the sentence-level diversity. 
 
\subsubsection{Informativeness Promotion}
Except for the diversity, 
we also evaluate the informativeness promotion with different normalization methods. 
We mainly take nouns and verbs into account to calculate the informativeness of a sentence. 
Moreover, we take the average sentence length into account to compare the informativeness, 
which is shown in Table~\ref{table:informativeness}. 
Even though Log Normalization barely promotes the diversity, 
it indeed promotes the frequency of verbs, especially on Subtitle dataset with an improvement of 12.1\%. 
With regard to the frequency of nouns, 
Log Mutual Normalization gets the best performance on all three datasets. 
As for the average sentence length, 
the conclusion comes that it almost correlates with the entity words (\emph{the total numbers of nouns and verbs}), 
which verifies that the sentence length indeed is able to indicate the informativeness of the sentence. 
Our normalization methods are able to promote not only  the diversity of generations 
but also the informativeness of generations. 
The main cause may be that our normalization methods reduce the probability of frequent but 
uninformative words, such as \emph{``the''}, \emph{``ok''} and \emph{``hello''}, 
and meanwhile increase the probability of less frequent but informative words, such as verbs and nouns. 

\subsubsection{Performance Promotion}
Furthermore, we compare the performance of models with different normalization methods 
from the aspects of BLEU and word entropy, shown in Table~\ref{table:bleu}. 
With regard to BLEU value, Mutual Normalization gets better performance. 
More specifically, Mutual Normalization improves BLEU by 29.2\% and 11.8\% on Subtitle and Lenovo respectively. 
However, Log Normalization gets a lower word entropy on both Subtitle and Lenovo. 
The reason mainly lies in that Mutual Normalization and Log Mutual Normalization make the 
words' frequency more uniform, leading to an increase of the word entropy.

Then we review the bigrams with frequency larger than 10 on Lenovo dataset 
and list several bigrams in Table~\ref{table:ngram-sample}. 
The first column indicates which word the bigram starts with. 
The numbers in Table~\ref{table:ngram-sample} indicates the diversity of bigrams, 
the larger the better. 
From the samples shown in the table, 
we can conclude that models augmented with our normalization methods 
indeed promote the diversity of bigrams. 
The last row indicates that Log Mutual Normalization 
gains the best performance in promoting diversity of bigrams.

According to Table~\ref{table:diversity}, \ref{table:informativeness} and \ref{table:bleu},  
it is interesting to see that the results on Subtitle corpus are more impressive than Twitter or Lenovo dataset 
for almost all evaluation metrics, including diversity, informativeness and BLEU. 
The main cause probably lies in that Subtitle corpus consists of sentence pairs acquired from ideal subtitles. 
However, Twitter and Lenovo corpus are more practical, 
which are made up of factual conversations between persons. 
In order to improve the loss function, we normalize $\hat{p}$ in Equation~\ref{equation:norm} 
with N-gram frequency calculated from wikipedia corpus. 
The most likely reason is that Wikipedia corpus and Subtitle corpus are more compatible.

\section{Conclusion} \label{section:conclusion}
In this paper, we propose three simple but efficient and effective data normalization methods 
to improve the dialogue generation. 
Our normalization methods are independent of generation models. 
Therefore, they are flexible and extensible to different generation models. 
By shrinking the margin of words' frequency to weaken the effect of Long Tail in linguistics, 
models augmented with our normalization methods 
are able to promote the diversity of generations at different levels. 
More importantly, enhanced models with our normalization methods are able to promote informativeness 
of generations to improve the performance. 
In the future, we will further improve the normalization methods with large real conversation corpus  
and verify the effectiveness of our normalization methods on other models.

\EOD

\end{document}